\documentclass[conference]{IEEEtran}
\IEEEoverridecommandlockouts
\usepackage{cite}
\usepackage{amsmath,amssymb,amsfonts}
\usepackage{algorithmic}
\usepackage{graphicx}
\usepackage{textcomp}
\usepackage{xcolor}
\def\BibTeX{{\rm B\kern-.05em{\sc i\kern-.025em b}\kern-.08em
    T\kern-.1667em\lower.7ex\hbox{E}\kern-.125emX}}
\begin{document}

\title{Comparing Machine Learning-Centered Approaches for Forecasting Language Patterns During Frustration in Early Childhood
}

\author{\IEEEauthorblockN{Arnav Bhakta}
\IEEEauthorblockA{\textit{Phillips Academy Andover}}
\and
\IEEEauthorblockN{Yeunjoo Kim}
\IEEEauthorblockA{\textit{Pennsylvania State University}}
\and
\IEEEauthorblockN{Pamela Cole}
\IEEEauthorblockA{\textit{Pennsylvania State University}}
}

\maketitle

\begin{abstract}
When faced with self-regulation challenges, children have been known the use their language to inhibit their emotions and behaviors. Yet, to date, there has been a critical lack of evidence regarding what patterns in their speech children use during these moments of frustration. In this paper, eXtreme Gradient Boosting, Random Forest, Long Short-Term Memory Recurrent Neural Networks, and Elastic Net Regression, have all been used to forecast these language patterns in children. Based on the results of a comparative analysis between these methods, the study reveals that when dealing with high-dimensional and dense data, with very irregular and abnormal distributions, as is the case with self-regulation patterns in children, decision tree-based algorithms are able to outperform traditional regression and neural network methods in their shortcomings.
\end{abstract}


\section{Introduction}
In child development literature, there is a longstanding view that the emergence of language skills in early childhood (ages 2 to 3 years) contributes to the development of self-regulation: children's ability to inhibit prepotent emotions and behaviors. Over the past 40 years, the only substantial evidence that has emerged to support this view is summarized by Cole et al. who found evidence of correlational associations between language development and emotional expressions; however, no direct causal links have been identified and it is unclear when and how language supports self-regulation of emotion \cite{b3}. 

In the current study, we seek to utilize information a data-driven machine learning (ML) approach provides to uncover patterns that theory may not suggest in how children use their language. Specifically, we employ 4 ML techniques – eXtreme Gradient Boosting (XGBoost), Random Forest (RF), Long Short-Term Memory (LSTM) Recurrent Neural Networks (RNN), and Elastic Net Regression (ENR) – to discover patterns in language use that can be used to forecast the effects of speech on children’s self-regulation of emotion using longitudinal observations of children during a frustration-invoking task at ages 2, 3, 4, and 5 years. The task was designed as follows: children were given one boring toy and a gift, and their mothers were given a questionnaire packet to complete. Mothers told their children they had to wait 8 minutes to open the gift until they finished their work. This task induced variable frustration among children as they found the blocked and delayed reward as problematic, making it possible to understand how children use language to express their interest and desire for the gift and frustration about waiting, at each age point.

Though this study is the first to use an ML-centered approach to describe when and how long children use speech in frustrated moments to examine the patterns of language use, similar algorithms have been used to provide accurate predictions on high-dimensional and dense data in a range of fields. For example, in comparing XGBoost and LSTM for predicting future retail products, it was found that XGBoost surpassed LSTM in accuracy \cite{b7}. Similarly, recent evidence suggested that RF outperformed LSTM in forecasting stock prices over both a 3-month and 3-year window \cite{b5}. Moreover, in predicting the spread of Creutzfeldt-Jakob Disease, ENR outperformed both RF and LSTM \cite{b1}. Hence, not only do we aim to characterize the frequency and duration of two different ways that children use their speech to verbally express their thoughts and feelings rather than the nonverbal emotion expressions (i.e., anger), but we also hope to compare the accuracy and effectiveness of each of the tested models and understand how well each is able to handle data with sudden and irregular bursts of speech, with no pre-established patterns.

The rest of this paper is organized as follows. Section 2 discusses the data set and pre-processing steps taken in this study. Section 3 provides preliminaries to deep learning and machine learning, before explaining the ENR, LSTM, and RF models that we use in this study. Section 4 discusses the formulation of the methodology used in the training of the models. Section 5 gives the results of each model as well as compares the accuracy of each model based on the methods highlighted above, with section 6 concluding the paper and talking about future opportunities that the work provides.

\begin{figure*}[ht]
    \centering
    \includegraphics[width=0.75\textwidth]{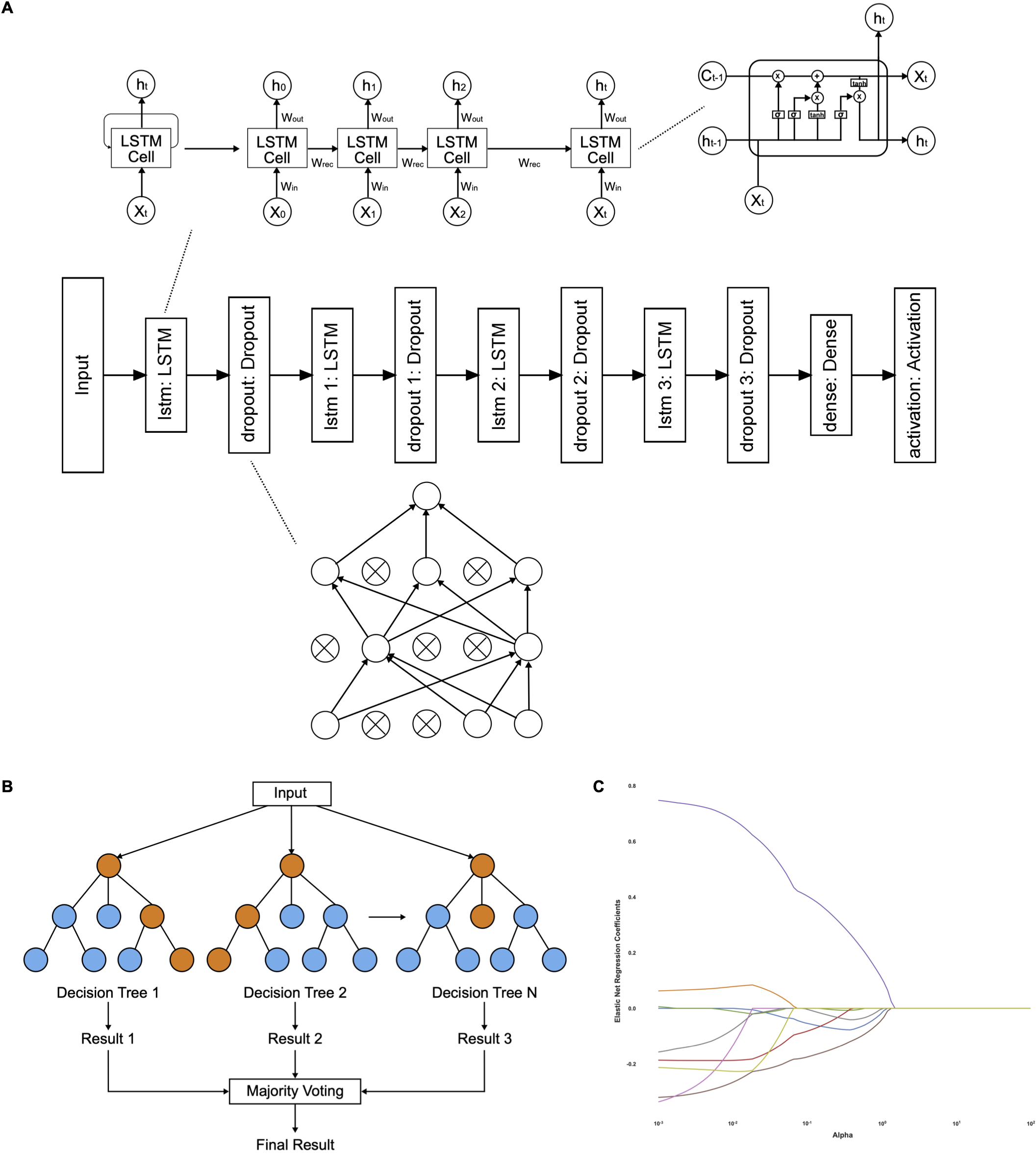}
    \caption{A visualization of the machine learning techniques used in this paper: A. LSTM Recurrent Neural Networks. B. Tree-Based Algorithms C. Elastic Net Regression}
\end{figure*}

\section{Dataset}

In the following section, we will discuss properties and characteristics of the dataset used in this study, and how the dataset was separated to differentiate between different categories of speech.

Throughout this work, we use archival data collected from a larger longitudinal study examining the socioemotional and language development in toddlers from economically strained families until they entered kindergarten \cite{b4}. Participants were recruited from rural and semi-rural households using U. S. Census data to target tracts with a higher volume of young families and a higher volume of families with incomes above the poverty threshold but below the median income. Recruiting focused largely on announcements from community leaders and community events, resulting in 124 families enrolling when their child was 18 months of age. Children were seen twice at this age (in-home and at the lab) and seen again every 6 months either at home or in the lab.

As previously described, each child participated in an 8-minute laboratory anger-eliciting task known as the wait task. Collecting second by second data of the children's utterances over the course of this task, made it possible for children's verbalizations to be coded via a binary coding system (0 = absence and 1 = presence) that categorized content of speech into utterances about the problem of waiting (e.g., “The toy is boring”, “I wanna open it [:gift]”, “When are you done?”) or utterances that reflect engaging in play behaviors or other thoughts irrelevant to the task (e.g., “Mom, what is this on my lip?”, “I love you”, “Is Allison at Nana’s house?”). 

By observing the frequency of speech at each age, it becomes apparent that the richest data lies within the age ranges of 3 to 5 years. As a result, we only use lab data from ages 3, 4, and 5 years in this paper. Additionally, for the purpose of this study, we categorize utterances about the problem of waiting as problem-related speech and utterances that reflect engaging in play behaviors or other thoughts irrelevant to the task as unrelated speech, through which we define two datasets for each age (a problem-related set and an unrelated set). In characterizing each child's problem-related and unrelated speech, it is important to note that they are highly variable in nature and often display sudden utterances after long periods of silence. Hence, we hope to compare how certain machine learning techniques are able to learn patterns in both the problem-related and unrelated sets at each age and predict frequencies of utterances across the 8-minute task for these two types of speech.

\section{Preliminaries}

In this section, we first discuss deep learning and neural networks, before providing an overview of Long Short-Term Memory Recurrent Neural Network Architectures, eXtreme Gradient Boosting, Random Forest, Elastic Net Regression, and the Ward Linkage Method.

\subsection{Long Short-Term Memory Recurrent Neural Network Architectures}

RNNs are a specific type of Deep Learning algorithm that have achieved high degrees of accuracy in pattern recognition and feature extraction. As opposed to Convolutional Neural Networks (CNNs) and Deep Neural Networks (DNNs) whose fixed-size vectors as inputs and fixed number of computational steps make it impossible for them to use prior outputs to better process and understand inputs in the current step, RNNs overcome this challenge by using a sequence of vectors over time that implements the outputs from the output units as inputs of the hidden layer:
\begin{equation}
\small
\begin{gathered}
h_t = \sigma_h( w_h x_t +u_h y^{t-1} + b_h)\\
y_t = \sigma_y( w_y h_t + b_y)
\end{gathered}
\end{equation}
Where $x_t$ is a vector of inputs, $h_t$ are hidden layer vectors, $y_t$ are output vectors, $w$ and $u$ are weight matrices, and $b$ is the bias vector.

Consequently, the loop RNNs utilize to pass data through the layers of the network, allows for greater pattern recognition and predictive modeling, which in cases such as ours, gives us the opportunity to both provide greater insight into the distribution of the data, but also compare the RNNS methodologies and accuracy to several other techniques \cite{b1}.

LSTMs are a subset of RNNs that serves to solve a recurring problem with current RNNs, being the vanishing gradient problem. In specific, in datasets with high dimensional and dense data, such as the one used in the current study, as more layers are added to the neural network, although a large change may occur in the input layer, a relatively small change is reflected in the output (see Fig. 1. A). Consequently, as more layers are added to the neural network, the backpropagation-derived gradients are continuously multiplied by the derivatives from the input layer to the output layer, leading to an exponential decrease in the gradient, in relation to the propagation of the input layer, and a much lower training accuracy, due to the weights of each layer not being updated correctly:
\begin{equation}
\small
\begin{gathered}
  \frac{\partial \varepsilon}{\partial \theta} = \sum_{1\leq t \leq T} \frac{\partial \varepsilon_t}{\partial \theta}\\
  \frac{\partial \varepsilon_t}{\partial \theta} = \sum_{1\leq k \leq t} \frac{\partial \varepsilon_t}{\partial x_t} \frac{\partial x_t}{\partial x_k} \frac{\partial^+ x_k}{\partial \theta}\\
  \frac{\partial x_t}{\partial x_k} = \prod_{t\geq i \geq k} \frac{\partial x_i}{\partial x_{i-1}} = \prod_{t\geq i \geq k} W_{rec}^{T} diag(\omega^{\prime} (x_{i-1}))
\end{gathered}
\end{equation}
Where $\varepsilon$ is the different outputs of the model, $x$ is the different inputs, and $W_{rec}$ represents the backpropagation algorithm \cite{b1}.

\begin{figure*}[ht]
    \centering
    \includegraphics[width=1\textwidth]{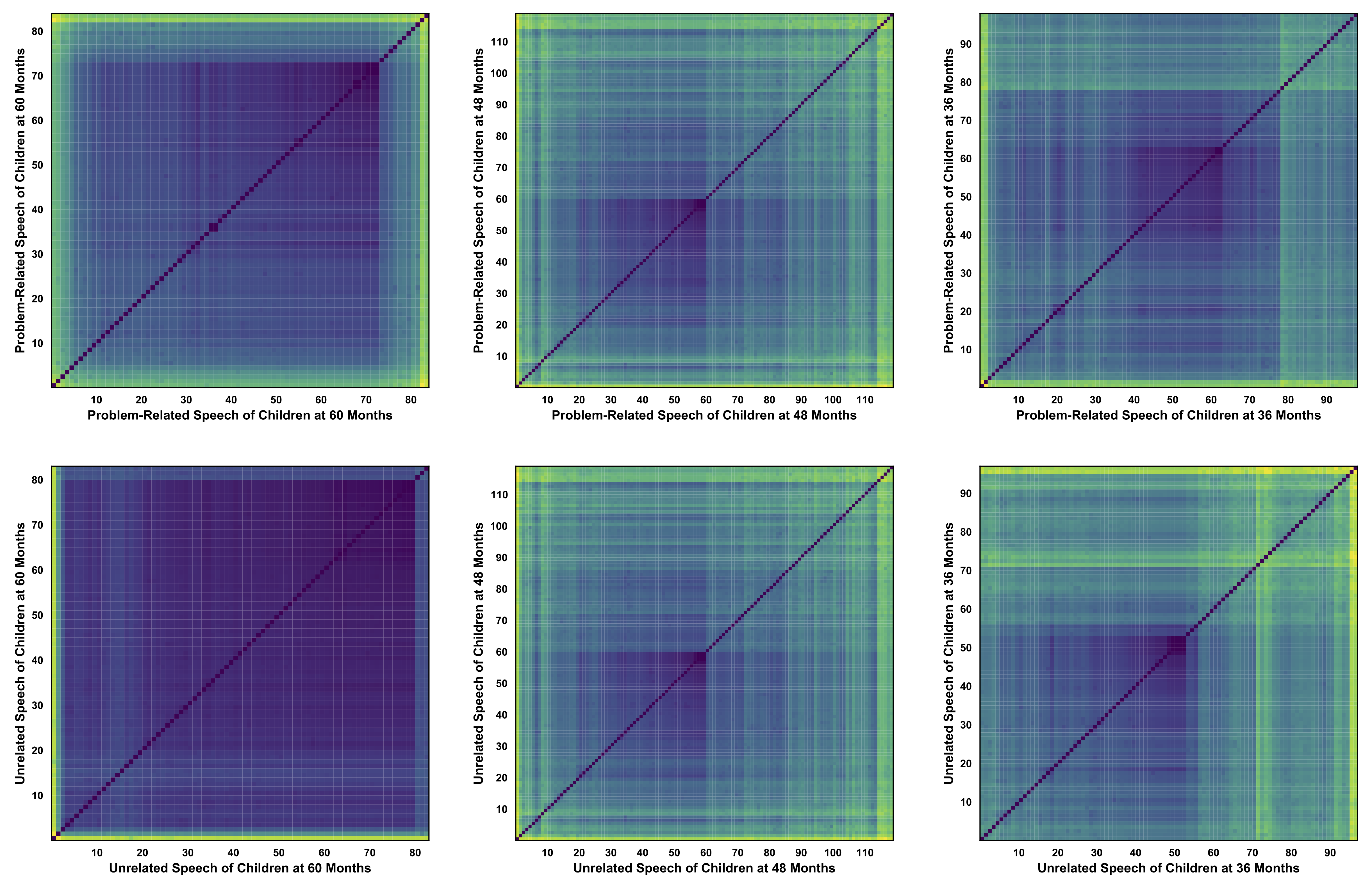}
    \caption{The data clustered via Ward's Method at 3, 4, and 5 years of age. A darker blue indicates a high degree of similarity between different children, while a light shade of yellow indicates a low degree of similarity between different children.}
\end{figure*}

LSTMs resolve this issue by maintaining a cell state for the model. As data flows from $C_{t-1}$ to $C_{t}$, the state is heavily regulated to minimize change and is only altered at 3 gates which are scomposed of a sigmoid neural net layer and a point wise multiplication operation, that is represented by the following algorithm:
\begin{equation}
\small
\begin{gathered}
  f_t = \sigma(W_f \times [h_{t-1}, x_{t}] + b_{f})\\
  i_t = \sigma( W_{i} \times [h_{t-1}, x_{t}] + b_{i})\\
  \tilde{C}_t = \tanh(W_{C} \times [h_{t-1}, x_{t}] + b_{C})\\
  \sigma_t = \sigma(W_{o} \times [h_{t-1}, x_{t}] + b_{o})\\
  h_t = o_t \times \tanh{C_t}
\end{gathered}
\end{equation}
Where $C_{t-1}$ and $C$ are the old and new cell states, respectively, $h_{t-1}$ and $x_{t}$ are the inputs, $\sigma$ is the sigmoid layer, $f_{t}$ is the input to the first gate, $i_t$ and $\tilde{C}_t$ and $i_t$ are the inputs to the second gate, $\sigma_t$ is the input to the third gate, and $h_t$ is the output.

Hence, because of an LSTM's ability to better handle high dimensional and dense data across several layers, we favor it instead of a traditional RNN. The basis of the LSTM model in this paper consists of 4 layers, through which the data flows to provide a final forecast, and 4 dropout layers in between the layers, to prevent overfitting \cite{b1}.

\subsection{Random Forest}

Random Forest (RF) is an ensemble of decision tree predictors capable of regression and classification tasks. Via the ensemble approach that RF utilizes, several individual classifiers are combined in one, to provide an accurate forecast on the training data. This is done by training the model on multiple decision trees with bootstrapping proceeded by aggregation, as opposed to a single, as depicted by Fig 1. B. By doing so, variance in the model is reduced, by training several unique decision trees on unique subsets of the features, as opposed to having a single input traverse down a single decision tree into smaller sets \cite{b1}.

The random portion of the algorithm stems from its allocation method. When splitting the input into unique subsets, it takes random subsets with replacement from the input and predictor variables at each node. It then calculates a weight for each input by carrying out a binary split at each node, using the predictor variable, and calculating the weighted average of all the terminal nodes. In following such a stringent procedure, RF provides accurate forecasts and a good generalization, due to the aggregation of singular decision trees \cite{b1}.

In the current study, we provide the training data to allow for a decision at each node, however, instead of treating the forecast as a regression task, we instead favor a classification approach to reduce error. Consequently, we hope to optimize the forecasts of the model.

\begin{figure*}[ht]
    \centering
    \includegraphics[width=1\textwidth]{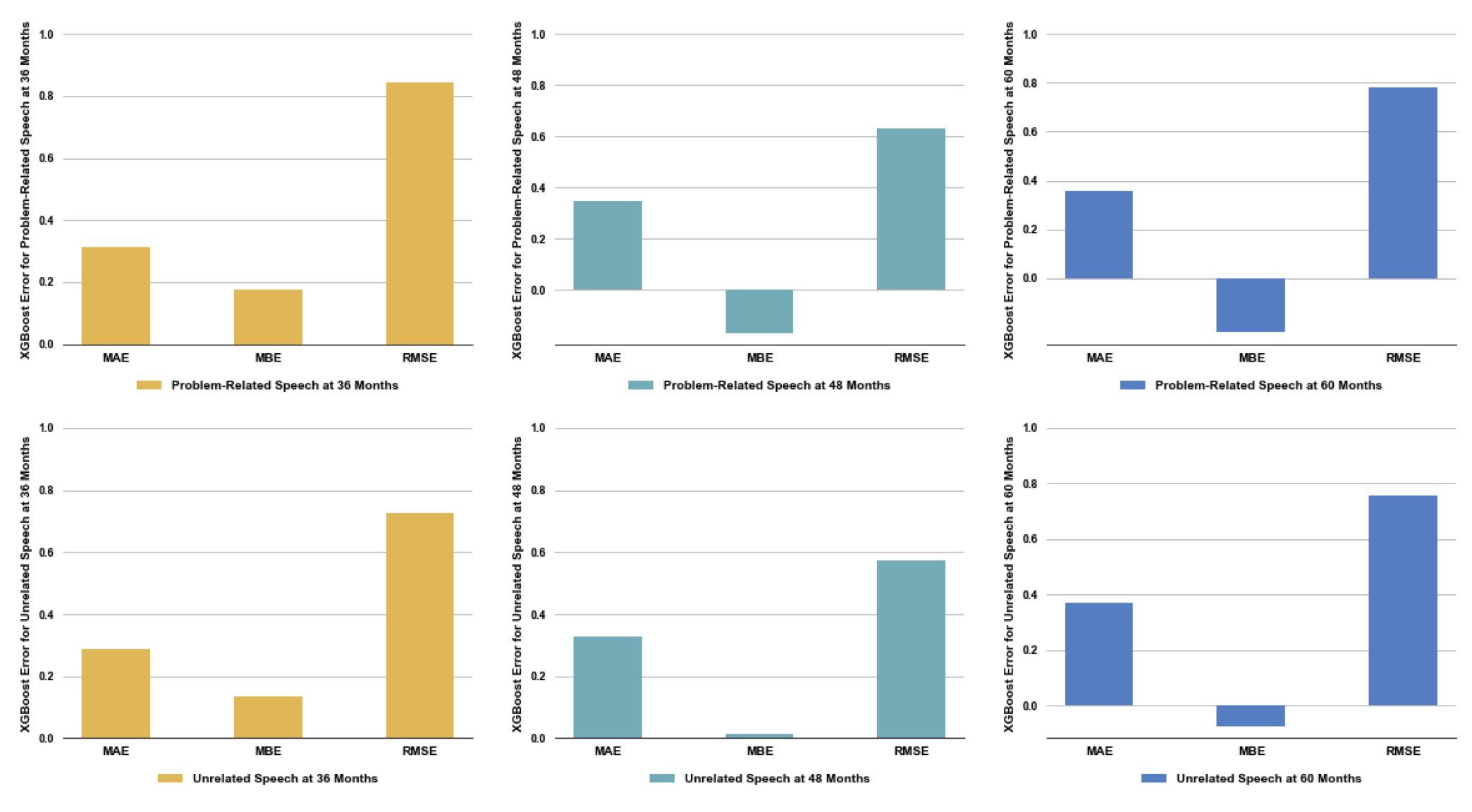}
    \caption{Results of forecasting the frequency of problem-related and unrelated speech at ages 3, 4, and 5 years using XGBoost.}
\end{figure*}

\subsection{eXtreme Gradient Boosting}

Similar to Random Forest, eXtreme Gradient Boosting (XGBoost) is a decision-tree-based ensemble algorithm that utilizes a gradient boosting framework (see Fig. 1. B for reference). By doing so, it is able to combine several independently weak classifiers, based on the gradient direction of the loss function, to create a single strong classifier with a higher degree of accuracy. Each independent classifier is based on random subsamples of the training set, generated with Bootstrap aggregation, that the model uses through several iterations. Additionally, the algorithm adds iterations of the model sequentially to adjust the weights of the weak classifiers, and consequently reduce bias and boost accuracy \cite{b2}. It does this by forecasting the predicted value, $\hat{y}_{i}$, using:
\begin{equation}
\small
  \hat{y}_{i} = \sum_{k=1}^{K}f_{k}(x_{i})
\end{equation}
Where $f_{k}$ us an independent decision tree and $f_{k}(x_{i})$ represents the prediction score given by that tree for the $i$-th sample \cite{b2}.

The set of $f_{k}$s is then used to minimize the objective function:
\begin{equation}
\small
  Obj = \sum_{i=1}^{n}l(y_{i}, \hat{y}_{i}) + \sum_{k=1}^{K}\Omega(f_{k})
\end{equation}
Where $l$ is the training loss function \cite{b2}.

The decision-tree-based ensemble model is then trained in an additive manner:
\begin{equation}
\small
\begin{gathered}
  Obj^{(t)} = \sum_{i=1}^{n}l(y_{i}, \hat{y}_{i}^{(t-1)}+f_{t}(x_{i})) + \Omega(f_{t})\\
  Obj^{(t)} = \sum_{i=1}^{n} [\partial_{\hat{y}_{i}^{(t-1)}}l(y_{i}, \hat{y}_{i}^{(t-1)})f_{t}(x_{i})\\+\frac{1}{2}\partial^{2}_{\hat{y}_{i}^{(t-1)}}l(y_{i}, \hat{y}_{i}^{(t-1)})f_{t}(x_{i})^2] + \Omega(f_{t})\\
\end{gathered}
\end{equation}
Where $\hat{y}_{i}^{(t)}$ is the the prediction of the $i$-th instance at the $t$-th iteration and $f_{t}$ is used to minimize the function. The second iteration of $Obj^{(t)}$ function is derived via a second order Taylor expansion \cite{b2}.

\begin{figure*}[ht]
    \centering
    \includegraphics[width=1\textwidth]{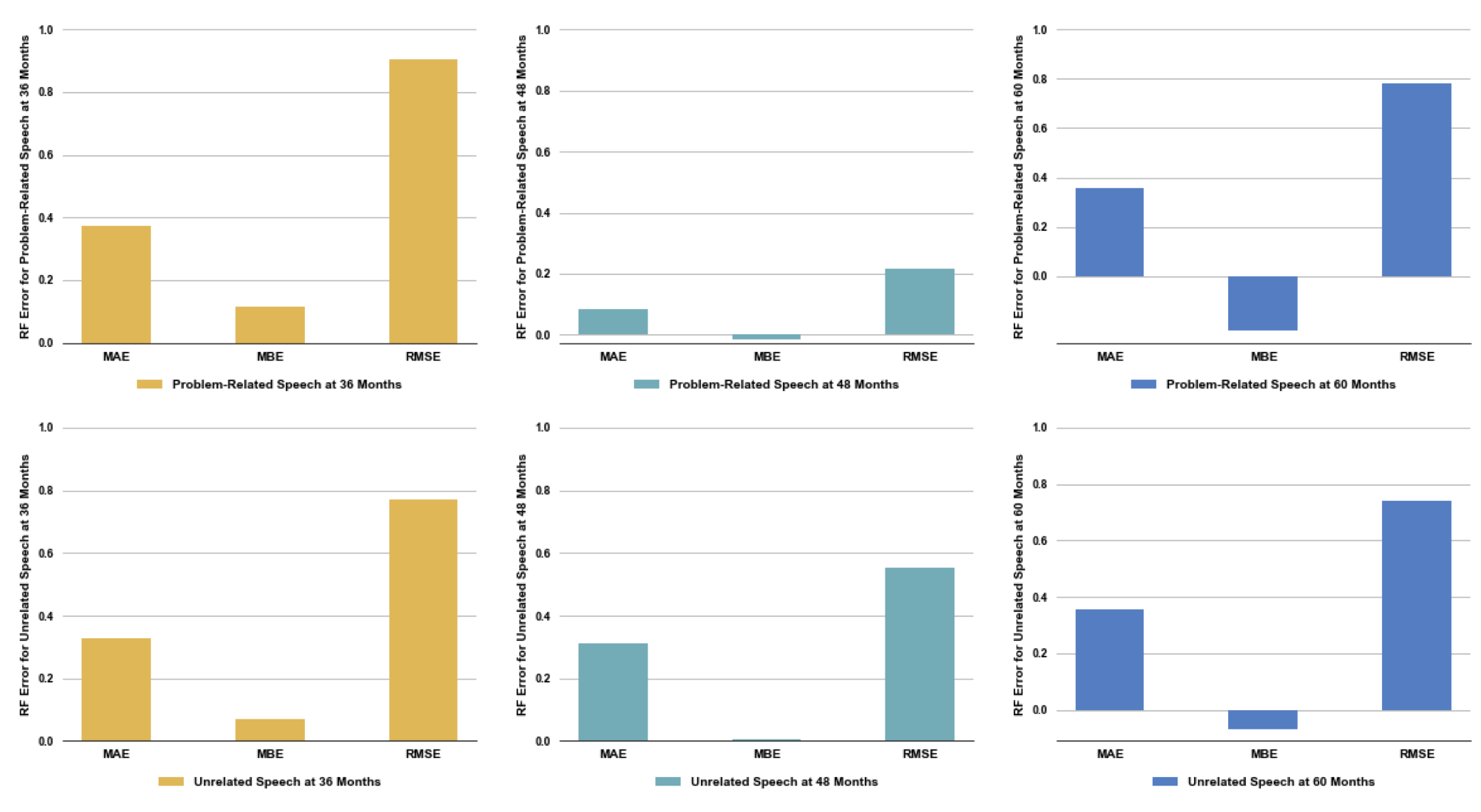}
    \caption{Results of forecasting the frequency of problem-related and unrelated speech at ages 3, 4, and 5 years using RF.}
\end{figure*}

One main way in which XGBoost builds off of and enhances the RF algorithm, is via the usage of a regularization term, $\Omega(f_{t})$, as seen in Eq (6), to prevent overfitting. This is done by tuning the hyperparameters and weighting feature's coefficients in accordance to their priority:
\begin{equation}
\small
  \Omega(f_{k}) = \gamma T + \frac{1}{2}\lambda \|w\|^{2}
\end{equation}
Where $\gamma$ and $\lambda$ are the regularization terms, $T$ is the number of leaves, and $w$ is the score at the respective leaf. Regularization will be discussed more in-depth in Section 3.4 \cite{b2}.

As with other decision-tree-based ensemble algorithms, while splitting a leaf node, the XGBoost models create several simple trees, to score it. XGBoost then uses a bagging classifier to use the prediction with the most votes and highest score to provide its final output forecast \cite{b2}.

Much like the RF model used, we approach the problem presented in this paper as a classification task in order in hopes of improving the accuracy of the forecasts, due to the lack of previous work with and information regarding the training set \cite{b2}.

\subsection{Elastic Net Regression}

With standard regression models a problem that arises with the inclusion of high dimensional and dense training sets, is variable selection due to relatively low sample sizes. Nonetheless, as is the case with XGBoost modeling, Elastic Net Regression (ENR) avoids this issue by generalizing many shrinkage-type regression methods into its model, namely Ridge and Lasso regression. Lasso Regression is attributed to an $\alpha = 1$ and utilizes an $l_1$ penalty to control parameter shrinking and variable selection. Ridge Regression is attributed to an $\alpha = 0$, but in contrast to Lasson regression, only utilizes an $l_2$ penalty on parameter shrinking. In generalizing both regression methods, ENR is able to attribute an $\alpha$ value between 0 and 1 ($0<\alpha<1$), accordingly letting the model assign a range of weights of automatic variable selection, as seen in Fig. 1. C. Doing so allows for variables that contribute more to prediction to be weighted heavily, and variables that contribute minimally to be weighted comparatively lower or not at all, enabling feature elimination and feature coefficient reduction. Such a generalization is carried out via the least angle regression algorithm, which tunes the penalty term to define the coefficients of features in a biased manner:

\begin{equation}
\small 
\lambda (\frac{1}{2}(1-\alpha)\beta^{2} + \alpha |\beta|)
\end{equation}

through the parameter $\alpha$, which has been generalized into the following equation \cite{b1}.

\begin{equation}
\small 
\hat{\beta}=\underset{\beta}{\operatorname{argmin}} \sum_{i=1}^N\left(y_i-\sum_{j=1}^p x_{ij} \beta_j\right)^2+\lambda_1 \sum_{j=1}^p|\beta_j|+\lambda_2 \sum_{j=1}^p \beta_j^2
\end{equation}

Where $\alpha$ controls the type of shrinkage, and the penalty parameter $\lambda$ controls the amount of shrinkage.

ENR also improves upon the issue of having correlated predictors in the training set, by subtracting a small number, $\epsilon$, from $\alpha$. By being incorporated into parameters shrinkage, the $\epsilon$ value allows the model to incorporate a greater number of correlated predictors. Thus, ENR models allow more leniency in the specifications of the training set needed to provide an adequate output.

We use ENR due to its ability to properly handle a high degree of similarity between predictors, as is the case in the current training set, and compare the accuracy of more traditional regression analyses in feature extraction and prediction, to more robust machine learning techniques \cite{b1}.

\begin{figure*}[ht]
    \centering
    \includegraphics[width=1\textwidth]{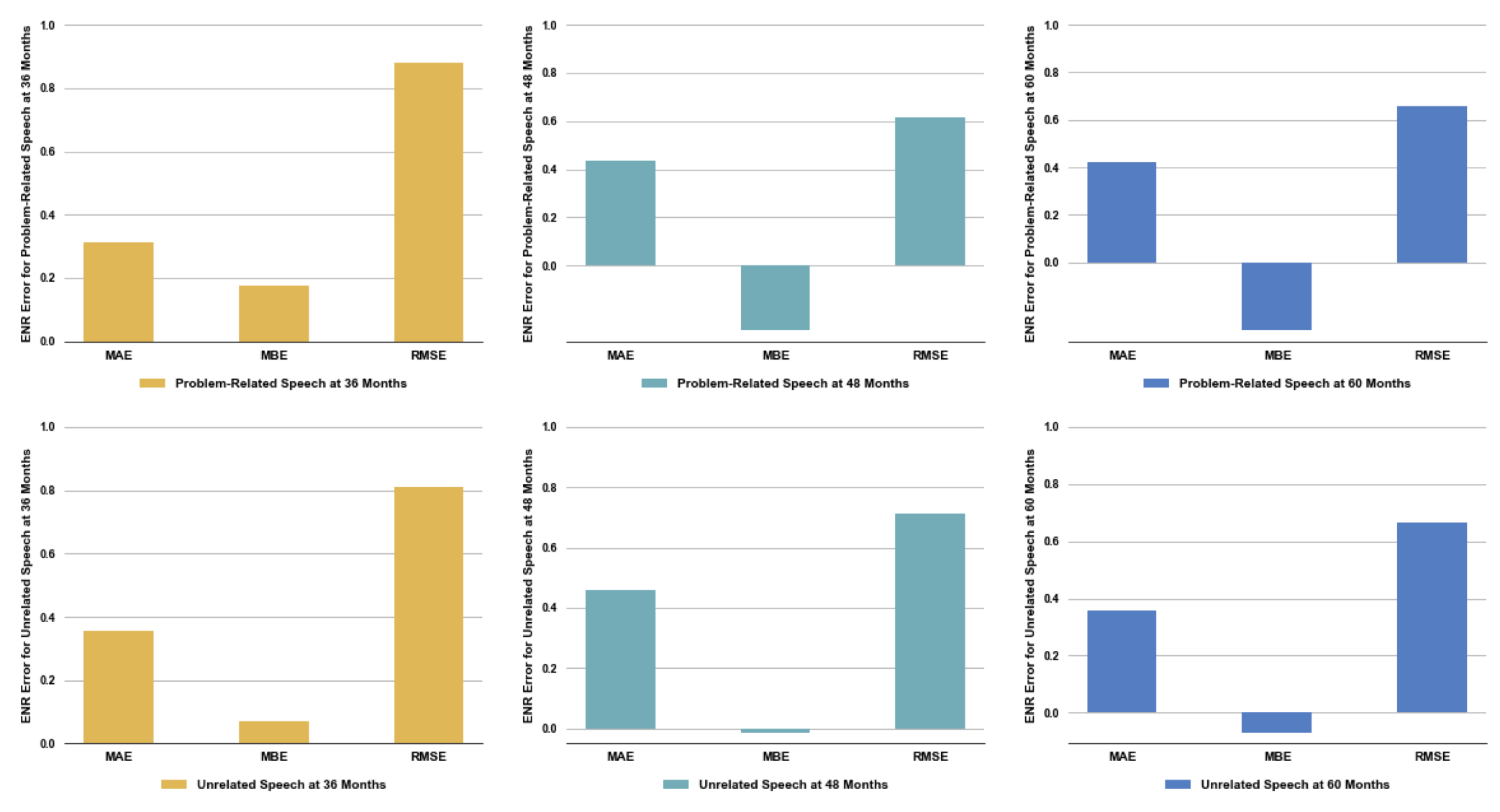}
    \caption{Results of forecasting the frequency of problem-related and unrelated speech at ages 3, 4, and 5 years using ENR.}
\end{figure*}

\subsection{Ward Linkage Method}

As mentioned in Section 3.3, due to the relatively low amount of work and information regarding the current training set, we employ hierarchical clustering to identify unique features of the set. In specific, we use the Ward Linkage Method, also known as Ward's Method, to cluster the set using a bottom-up approach or an agglomerative approach, wherein each point in the set starts in its own cluster, and is then joined greedily with other clusters until an equilibrium point is reached at which no more clusters can be merged. In our case, each point is considered to be a single time series in the training set. Ward's Method determines whether to merge two clusters based on the variance between the clusters \cite{b6}. It determines variance by defining it as the distance between two clusters, A and B, in space:
\begin{equation}
\small
\begin{gathered}
  \Delta(A, B) = \sum_{i \in A \cup B} \|\Vec{x}_{i} - \Vec{m}_{A \cup B}\|^{2} - \sum_{i \in A} \|\Vec{x}_{i}-\Vec{m}_{A}\|^{2}\\ - \sum_{i \in B} \|\Vec{x}_{i}-\Vec{m}_{b}\|^{2}\\
  \Delta(A, B) = \frac{n_{A}n_{B}}{n_{A}+n_{B}}\|\Vec{m}_{A}-\Vec{m}_{B}\|^{2}
\end{gathered}
\end{equation}
Where $\Delta(A, B)$ is the merging cost of combining clusters A and B, $\Vec{m}_{j}$ is the center of cluster $j$, and $n_{j}$ is the number it points to \cite{b6}.

As each point begins with its own unique cluster in space, $\Delta(A, B)$ initially starts with a value of 0, but grows upon the merging of different clusters. Ward's Methods aims to minimize this growth by finding the smallest ordinary distance between two points, p and q, in Euclidean Space:
\begin{equation}
\small
d(q,p) = \sqrt{\sum_{i=1}(q_{i}-p_{i})^2}
\end{equation}

The algorithm then determines if merging with more clusters is necessary using the Calinski and Harabasz (CH) Index to evaluate the possibility of segmentation based on within and between cluster variance \cite{b6}. By minimizing the within-cluster variance and maximizing the between-cluster variance, it chooses a number of clusters, K, that maximized the CH score:
\begin{equation}
\small
CH(K) = \frac{B(K)/(K-1)}{W(K)/(n-K)}
\end{equation}
Where W is the within-cluster variation and B is the between-cluster variation.

In this paper, we favor Ward's Method due to its ability to create compact and evenly-sized clusters, and clearly define individual time series in the training set that can be separated into a subset for which to base training on \cite{b6}.

\begin{figure*}[ht]
    \centering
    \includegraphics[width=1\textwidth]{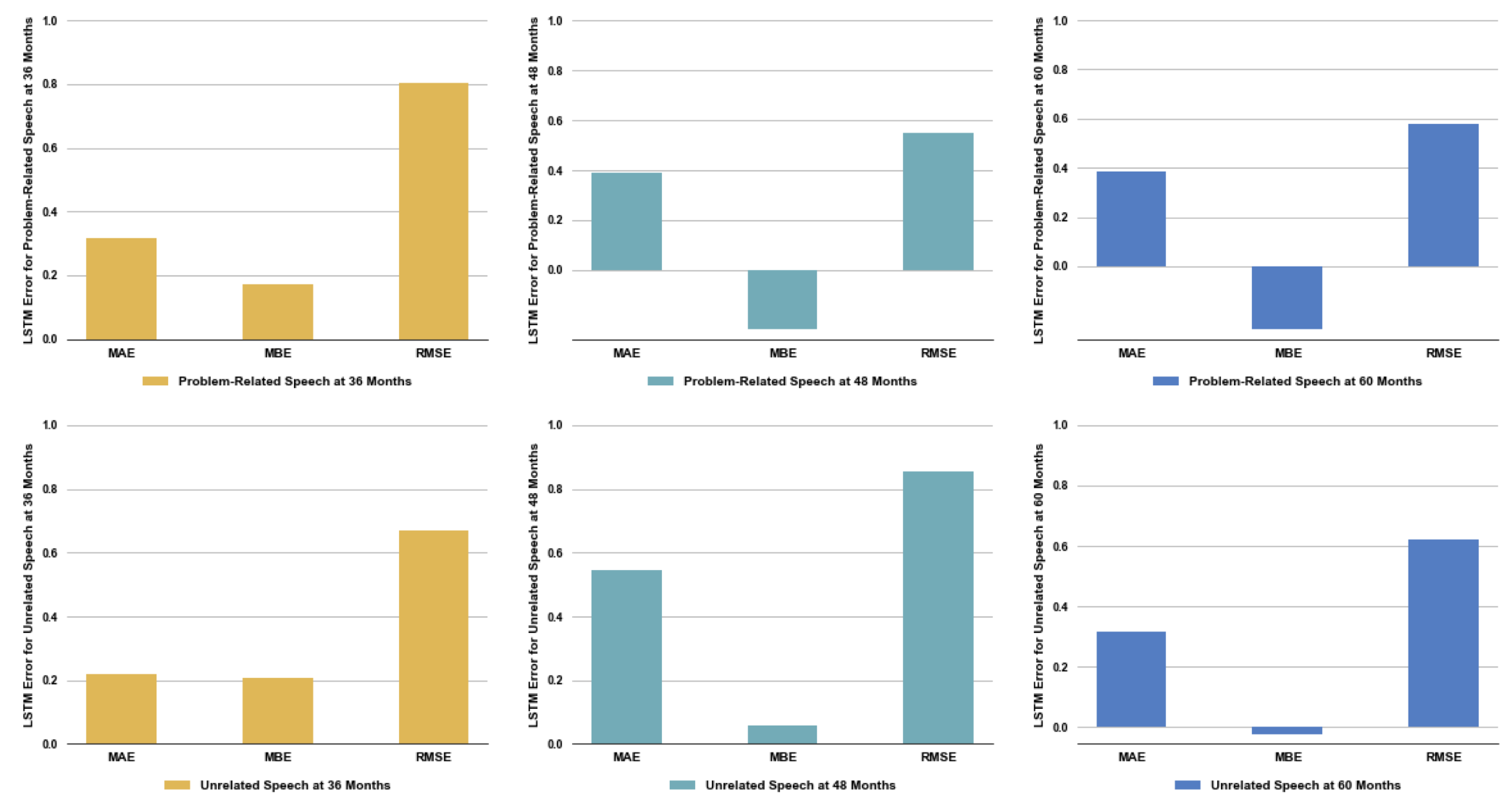}
    \caption{Results of forecasting the frequency of problem-related and unrelated speech at ages 3, 4, and 5 years using LSTM.}
\end{figure*}

\section{Formulation}

In formulating how to approach the task at hand, it is vital that we realize that there is not necessarily a defined set of predictors and responses for which we can use as a basis for training. Consequently, in an effort to define such a set, we employ Ward's Method to cluster the problem-related and unrelated speech at each age (Fig. 2). As can be observed, across all ages, there is a high degree of similarity between each of the children that make up the middle portion of the set, while at the edge cases, there is a high degree of dissimilarity. Accordingly, in defining our predictor and response sets, we attribute all edge cases that are indicated to have high degrees of dissimilarity to the response set, and the remaining cases to the predictor set, due to the fact that we want to avoid high degrees of multicollinearity between our response and predictor set and consequently, remove the risk of having a reduced precision in predicting weights of each model.

These sets are then split into a training and testing split, using a 70/30 split, respectively, where the training data is used to construct the model, while the testing data tests the validity of the trained model. Additionally, before training the model, the data is normalized to ensure consistency in the dataset and that each data type has the same format.

Prior to training, the training set is pre-trained on an ordinary least-squares (OLS) model or a simple linear regression model, of which half of this set is also pre-trained on a Vector Auto Regression (VAR) model, in an effort to smooth the data, before it is passed in as inputs via which to fit all of the specified models. To evaluate the effectiveness of each model in predicting the frequencies of problem-related and unrelated speech at each age, we use 3 methods to measure the error in the model's predicted frequencies on the test sets. These methods include Root Mean Square Error (RMSE), Mean Bias Error (MBE), and Mean Absolute Error (MAE).

\section{Results}

In this section, we study how well each of the proposed models was able to forecast the frequency of problem-related and unrelated speech at each of the 3, 4, and 5 years of age. 

\subsection{eXtreme Gradient Boosting}
The results of the XGBoost model are shown in Fig. 3. We observe that our XGBoost model performs well at all ages, achieving a minimum RMSE of 0.629 when forecasting problem-related speech at 4 years of age, while it achieves a maximum RMSE of 0.843 when forecasting problem-related speech at 3 years of age. Moreover, it received relatively low MAEs as well, with a minimum of 0.287 being reached when predicting the frequency of unrelated speech at 3 years of age, and a maximum of 0.367 being reached when predicting the frequency of problem-related speech at 5 years of age, once again suggesting that it was quite accurate in forecasting different frequencies of problem-related and unrelated speech across all ages, as the frequency in speech could range from 0 seconds to 36 seconds. 

However, 50\% of the model's MBEs were less than 0, being when it forecasted on unrelated speech at 3 years of age (MBE = -0.170), problem-related speech at 5 years of age (MBE = -0.219), and unrelated speech at 5 years of age (MBE = -0.076), indicating that the model had a tendency to overfit its predictions for these specific cases. 

\subsection{Random Forest}
The results of the RF model are shown in Fig. 4. Our RF model performs most optimally out of all of the tested models, achieving a minimum RMSE of 0.217 when predicting the frequency of problem-related speech at 4 years of age, and a maximum RMSE of 0.904 when predicting the frequency of problem-related speech at 3 years of age. Similar to the XGBoost model, the RF model receives its maximum and minimum RMSEs when forecasting using the same respective ages and language types as the XGBoost model. Comparatively though, it achieves an average RMSE of 0.599 which is much lower than that of the XGBoost model, which receives an average RMSE of 0.718. Moreover, although it achieves its minimum and maximum MAEs while forecasting on different sets than the XGBoost model, being when forecasting on problem-related speech at 4 years of age (MAE = 0.086) and on problem-related speech at 3 years of age (MAE = 0.373), it once again receives an average MAE lower than that of the XGBoost model, being 0.300 as opposed to 0.317.

In addition, 50\% of the RF model's MBE scores were less than 0, occurring at the same ages and speech types as that of the XGBoost model, indicating that though generally, the RF model outperformed the XGBoost model in accuracy, the both still overfit for the same cases. 

\subsection{Elastic Net Regression}
The results of the ENR model are shown in Fig. 5. Our ENR model received an average RMSE of 0.724, with its minimum RMSE occurring when forecasting problem-related speech at 4 years of age, for which it received an RMSE of 0.617, and its maximum RMSE occurring when forecasting on problem-related speech at 3 years of age, for which it received an RMSE of 0.880. Additionally, it achieved a minimum MAE of 0.312 when predicting the frequency of problem-related speech at 3 years of age, and a maximum MAE of 0.460 when predicting the frequency of unrelated speech at 4 years of age, reaching an average MAE of 0.3905.

When comparing the results of the ENR model to those of both decision tree-based models, it is evident that ENR performed worse in both metrics. Interestingly though, unlike the RF and XGBoost models, which both performed their best and worse when forecasting on the same age and language types, the ENR model, struggled and performed optimally when dealing with different cases, suggesting that the methodology used by decision tree-based algorithms may be more optimal than a more traditional regression-based algorithm. This point is further supported by the fact that 75\% of the ENR model's predictions received MBEs less than 0, pointing out that it is more likely to overfit the data than both the decision tree-based approaches.

\subsection{Long Short-Term Memory}
The results of our LSTM model are shown in Fig. 6. Our LSTM model achieved a minimum RMSE of 0.546 when predicting the frequency of problem-related speech at 4 years of age, and a maximum RMSE of 0.855 when predicting the frequency of unrelated speech at 4 years of age, allowing it to receive an average RMSE of 0.679. Furthermore, the model received its minimum MAE of 0.219 when forecasting on unrelated speech at 3 years of age, and its maximum MAE of 0.544 when forecasting on unrelated speech at 4 years of age. Similar to the decision tree-based models, our LSTM model had an MBE less than 0, 50\% of the time, with these occurrences being at the same positions at which the XGBoost and RF models also had MBEs less than 0.

In analyzing the implications of these results, it is clear that RF provides more accurate predictions and pattern recognition than LSTM when forecasting on similarly characterized data as being used in the current study. Moreover, although both models are likely to overfit when dealing with similar data and cases, the RF model faces the highest degree of difficulty when dealing with a different subset of the test set than the LSTMs, highlighting how when fitting data similar to the test sets, the decision tree framework of RF allows it to better handle the data than LSTMs, who often come up short in these instances.

\section{Conclusion}
In this paper, we investigate how accurately 4 different models – eXtreme Gradient Boosting, Random Forest, Long Short-Term Memory Recurrent Neural Networks, and Elastic Net Regression – are able to predict the frequency of problem-related and unrelated speech at 3 separate ages: 3, 4, and 5 years of age. In doing so, we observed that decision tree-based methodologies were consistently able to outperform more traditional regression-based algorithms in identifying patterns in language, and forecasting based on them. Moreover, we discovered that when working with high-dimensional and dense data, that follows a very sudden and arbitrary trajectory, Random Forest greatly outperformed eXtreme Gradient Boosting, while Long Short-Term Memory Recurrent Neural Networks also provided more accurate predictions than eXtreme Gradient Boosting models, but fell short to Random Forest models. 

Furthermore, in analyzing in which cases the decision tree-based models outperformed the other tested models, we found that while both eXtreme Gradient Boosting and Random Forest performed their worst and best on the same test cases, this was not the case for the other tested models, as the decision tree frameworks were properly able to handle the data that the other models faced difficulties with, suggesting that when dealing with similarly characterized data, decision tree-centered algorithms are better suited for pattern recognition and forecasting based on these observations. Additionally, this paper is the first to use a machine learning approach to describe when and for how long children use two different types of verbalizations during a frustrating task. The patterns indicate that at age 3 years children refer to the demands of waiting throughout the task whereas, by age 5 years, children mainly verbalize these concerns at the outset of the 8-minute wait and, again, toward the end of the task period. This suggests that by age 5 years the children are regulating the frustration of the wait at least during the middle period of the wait. This has not been predicted by theory and points out that by age 5 years it is not that the children no longer mind waiting but that they are coping it with differently, perhaps in a more regulated way, by the time they reach age 5 years. This new information lays a base for future endeavors with integrating machine learning to gain a deeper view of how children use their language when their desires are frustrated and how the use of language during such a situation changes with age.

\end{document}